# Exploring Human Crowd Patterns and Categorization in Video Footage for Enhanced Security and Surveillance using Computer Vision and Machine Learning

**Afnan Alazbah[1], Khalid Fakeeh[2], Osama Rabie[3]**

[1]aazbah0001@stu.kau.edu.sa, [2]kfakeeh@kau.edu.sa, [3]obrabie@kau.edu.sa

[1, 2, 3]Information Systems Department, King Abdul-Aziz University, Jeddah, Saudi Arabia

*Abstract*— **Computer vision and machine learning have brought revolutionary shifts in perception for researchers, scientists, and the general populace. Once thought to be unattainable, these technologies have achieved the seemingly impossible. Their exceptional applications in diverse fields like security, agriculture, and education are a testament to their impact. However, the full potential of computer vision remains untapped. This paper explores computer vision's potential in security and surveillance, presenting a novel approach to track motion in videos. By categorizing motion into Arcs, Lanes, Converging/Diverging, and Random/Block motions using Motion Information Images and Blockwise dominant motion data, the paper examines different optical flow techniques, CNN models, and machine learning models. Successfully achieving its objectives with promising accuracy, the results can train anomaly-detection models, provide behavioral insights based on motion, and enhance scene comprehension.**

## I. INTRODUCTION

The advancement of technology, including artificial intelligence and robotics, has significantly bolstered the realm of security and surveillance. The amalgamation of machine learning techniques with surveillance practices has emerged as a potent solution to address issues such as crime, illicit activities, and even violent protests. Recent experiences have underscored the value and necessity of automated video surveillance. Thanks to computer vision, a technology that enables computers to interpret visual information like humans, tasks like identifying people in frames, tallying individuals in crowded scenes, spotting unusual behaviors, and analyzing motion in surveillance videos are carried out without human intervention. However, the analysis of crowd motion and detection of abnormal behaviors has always been a challenging endeavor due to the numerous factors influencing individual movement. By deciphering crowd motion, potential incidents of violence, riots, traffic congestion, and stampedes can be averted. In the general context and motivation behind this study, as referenced in [1], the primary goals of automated surveillance video analysis encompass continuous monitoring, reducing labor-intensive tasks, recognizing objects or actions, and comprehending crowd behavior. This paper delves into the identification of diverse forms of crowd motion and tracking abnormal behaviors utilizing Convolutional Neural Networks (CNN). Most of the research in crowd analysis is concentrated on the following aspects: counting the crowd, categorizing crowd types based on density, detecting motion within frames, and classifying various types of motion. Crowd counting plays a pivotal role in ensuring safety and security. It facilitates event planning, traffic management, and gauging capacity in various situations. However, counting densely packed crowds poses a considerable challenge. Research, such as [2], indicates that over 17% of papers on crowd analysis focus on crowd counting. For instance, [3] presents a comprehensive exploration of crowd counting types and diverse algorithms, proposing a novel method utilizing the statistics of spatio-temporal wavelet sub-bands. Another approach [4] employs multiple sources and Markov Random Fields to count individuals in dense crowds. Categorizing crowd types based on density is essential for comprehending the dynamics of motion. According to [5], crowds can be classified into three types: microscopic, mesoscopic, and macroscopic, dependent on crowd density. Microscopic view entails understanding the flow of individuals within a limited frame, while mesoscopic view accounts for a larger crowd, and macroscopic view involves a densely filled frame. The forces at play in these scenarios differ, driving crowd motion. Detecting motion within frames can be achieved by training a model using a CNN architecture or by tracking individual points using optical flow. For instance, [6] utilized Shi-Tomasi Corner Detection and the Lucas–Kanade algorithm to detect crowd motion. Similarly, [7] employed Motion Information Images (MII) to train a CNN model for motion and abnormality detection. Identifying types of motion is invaluable for understanding crowd behavior, event planning, mitigating traffic congestion, and anticipating abnormal motion. Researchers like [8] utilized VGG16 CNN architecture models to classify crowd types as homogenous, heterogeneous, or violent. Additionally, [9] used mathematical concepts like Taylor's theorem and Jacobian matrix to categorize crowd motion into five generic types: Lanes, Arc/Circle, Fountainheads, Bottlenecks, and Blocks. This paper endeavors to classify various crowd motion types, such as Arcs, Lanes, Converging/Diverging, and Blocks/Random, with a focus on drone footage analysis employing cutting-edge CNN and machine learning techniques. The paper also delves into different optical flow methods, examining their pros and cons. The proposed model aids in understanding scene motion and supports the training of multiple anomaly detection techniques. The paper's objectives include identifying key features in frames for motion tracking, exploring noise reduction options, devising an improved point tracking approach, generating data for anomaly detection, creating datasets for CNN and machine learning models, comparing methodologies, and enhancing the model's performance for dynamic drone footage analysis in security scenarios.

The paper is as follows in the next section we will see the background related to our paper. In Section III, the works that are related to our technique is been presented. In Section IV, the dataset description with preprocessing are presented. In Section V, the experimental approach is been described. In Section VI, the discussion of results are presented and we conclude the paper in Section VII with some conclusion and future works.

## II. BACKGROUND

Computer vision has developed from a multitude of complex theories, algorithms, and models. The primary focus of this paper revolves around video surveillance. This section aids in comprehending the specific technical complexities relevant to this domain.

### A. Optical Flow

Optical flow, which can be regarded as one of the fundamental concepts in the realm of computer vision, plays a pivotal role in deciphering the complex patterns of object movement from one frame to another. This concept finds widespread applications in various fields such as robotics, image processing, motion detection, and object segmentation. In essence, optical flow allows us to uncover the motion patterns hidden within consecutive frames of videos, providing crucial insights into the dynamics of objects over time. When we think of videos, we envision a sequence of images, each potentially unrelated to the others. However, in the real-time scenario, a video captures the sequential changes in pixel values over a specific period. This dynamic nature of videos leads us to explore algorithms that unveil the relationships between pixels in different frames. In a recent study [10], various optical flow algorithms were meticulously examined and evaluated. The findings of this study pinpoint the Lucas-Kanade Algorithm as the most promising one among the eight algorithms assessed. Visualizing optical flow is often represented through diagrams using vectors that indicate the changes from one frame (let's call it Frame F1) to another (Frame F2). Yet, in real-time analysis, it's practical to focus on specific points that yield more meaningful insights. To illustrate, consider the movement of a hand from Frame F1 to Frame F2. This movement might involve changes in hundreds of pixels, many of which could be redundant. Rather than analyzing the entire hand movement, it's more informative to concentrate on the flow of pixels at the corners of the hand.

This is where corner detection algorithms come into play. These algorithms are designed to simplify the complexity of analysis while enhancing the performance of optical flow algorithms. Within the context of corner detection, this paper delves into two distinct techniques to determine the optimal approach for the task at hand. The first is the Shi-Tomasi Corner detection method, which bears resemblance to the Harris Corner Detector. This technique is widely employed to identify interest points and feature descriptors in images. Interest points can take the form of corners, edges, and blobs. Importantly, these interest points remain invariant in the face of rotation, translation, intensity changes, and scaling variations. The differentiating factor between Harris and Shi-Tomasi corner detection lies in the computed R value, which is used to identify corners. On the other hand, the Features from Accelerated Segment Test (FAST) Corner detection technique employs an alternative strategy. It predicts not only corners but also edges based on color intensity and a specific threshold.

### B. Density-Based Clustering

Clustering, in simple terms, involves grouping together objects that are alike. This similarity is determined by factors such as shape, angle, size, and position. To make the best use of a computer's memory, which is like its thinking space (CPU/GPU), it's crucial to focus on the most important points for analysis. So, to handle this, we gather the points and arrows according to where they are and where they're headed. This helps us bring similar points and arrows together, which then lets us figure out how a crowd might move. In our paper, we've looked at two kinds of density-based clustering techniques such as DBSCAN and OPTICS.

### C. Convolutional Neural Networks

CNNs are an advanced concept within the realm of neural networks, granting computers the ability to comprehend images and videos. These networks have found extensive applications across various fields, including robotics, face detection, crowd detection, weather analysis, advertising, and environmental studies. At the core of CNNs are individual units called neurons, each equipped with learnable weights. These weights, initially set randomly, can be adjusted through training to develop a model that accurately performs tasks. CNNs consist of three main components: Convolutional Networks (ConvNets), pooling layers, and fully connected layers. ConvNets, also known as convolutional layers, play a crucial role in altering the pixel values of an image through the use of filters. Think of an image as a grid of pixels, and these filters as templates that help modify pixel values using matrix multiplication. These filters are applied across the entire image through a process called striding. Pooling, another essential concept, involves reducing the image's size with the help of specialized filters. There are two primary types of pooling: average pooling and max pooling. In average pooling, the filter computes the average pixel value within a specified region, while in max pooling, only the maximum pixel value is retained. This reduction in image size aids in maintaining important features while reducing computational complexity. Fully connected layers, the third component, resemble traditional neural networks. They take the 1D arrays generated by the ConvNets as inputs and consist of various hidden layers that are interconnected. The outputs of these layers serve as classification nodes, potentially representing integer labels or one-hot encoded values that predict the class of the input. The SoftMax function is often used to make predictions by selecting the node with the highest probability. Several established CNN architectures are available as modular components. These pre-trained modules can be directly implemented if the inputs and outputs match the module's expectations. The PyTorch library offers a platform to explore and utilize these networks effectively.

This paper focuses on the implementation of three specific CNN architectures: AlexNet, VGG, and ResNet. AlexNet, proposed by Alex Krizhevsky in 2012 [11], marked a significant breakthrough. Its architecture comprises 8 layers, with 5 convolutional layers followed by max pooling, and the final 3 layers being fully connected. Notably, AlexNet utilized the non-saturating ReLU function, which can be replaced with tanh or sigmoid functions for enhanced performance. VGG, introduced by Karen Simonyan in 2015 [12], offers variations based on network depth: VGG11, VGG13, VGG16, and VGG19. This architecture takes a 224*224*3 image as input and subjects it to several convolutional layers followed by max pooling. This paper specifically implements the VGG11 architecture, which deviates from AlexNet's design by altering the placement of max pool layers. In VGG, these layers may appear after a series of 2 or 3 convolutional layers. This strategy helps retain feature information before size reduction via max pooling. Towards the end of the architecture, there are three fully connected layers for classification purposes. ResNet, short for residual neural network, draws inspiration from the brain's pyramidal cells [13]. It utilizes skip connections to allow data to pass through certain layers without modification. The model employs double or triple skips using ReLU activation and batch normalization. The foundational idea behind ResNets is that adding more layers to a CNN won't necessarily reduce errors, but it could increase computational costs. Thus, ResNets propose using skip connections to efficiently reach lower layers when needed, preserving essential feature characteristics.

### D. Machine Learning Models

Machine learning involves different approaches to solving problems. There are three primary types of machine learning algorithms: supervised learning, unsupervised learning, and reinforcement learning. Supervised learning is like learning from past experiences with a bit of guidance. In this approach, models are trained using labeled data, which means data with clear indications of what they represent. The model learns from these labeled examples to make predictions on new, unseen data. The goal here is to check how accurately the model can predict the right outcomes. Unsupervised learning is a bit different. Imagine learning from a bunch of unsorted notes without prior context. Here, the models work on raw data without any labels or indications. These models are fed large amounts of data and they try to uncover patterns, similarities, and structures within the data itself. This helps in understanding relationships and hidden insights within the information. Reinforcement learning, on the other hand, is like learning by trial and error. It's as if a machine is placed in an environment and learns by continuously experimenting and adapting based on the outcomes of those experiments. This approach is often used in training machines to make sequential decisions, like training a robot to navigate through a complex maze. This paper mainly deals with classification models, which are designed to categorize data into specific groups. In the realm of supervised learning, various models are used for classification, each with its own way of making sense of the data. Logistic regression, for instance, estimates discrete values based on the input data. It tries to find a relationship between the input variables and a binary outcome, like whether an email is spam or not. But when you're dealing with more than two categories, you step into the territory of multinomial logistic regression. In this scenario, a linear function is not suitable for the best fit. Instead, the sigmoid function is employed to make predictions as demonstrated in the earlier section. Support Vector Machines (SVM) are like teachers who learn from examples. They plot each data point in a multi-dimensional space and aim to draw a line or boundary that best separates the different classes of data. SVMs are quite effective in classification tasks, especially when the data isn't easily separable. K-Nearest Neighbours (KNN), on the other hand, takes the "birds of a feather flock together" approach. It groups data points based on their proximity in the feature space. The "k" in KNN refers to how many neighboring data points you consider when making a classification decision. Gaussian Naive Bayes is a method that applies the Bayes' theorem. It's like making predictions based on known probabilities and the independence of different factors. It's quite simple yet effective, especially for text-based classifications like email spam detection. Another technique such as Perceptron is like a building block of a simple neural network. It learns to classify data that is linearly separable, which means data that can be separated using a straight line. It's like learning to draw a line that best divides different types of objects. Stochastic Gradient Descent (SGD) is a variation of a well-known optimization technique called Gradient Descent. It's like a more efficient way of figuring out the best fit of a model to the data. Instead of using all the data to adjust the model's parameters, it uses small batches, which speeds up the process, especially for large datasets. Decision Trees are like making a series of yes-or-no questions to classify data. It's as if you're asking a series of questions to reach a conclusion. Each node in the tree represents a question or a decision, and the branches represent the possible answers or outcomes. Random Forest is like a committee of decision trees, each with their own opinions. Each tree in the forest gets a vote, and the majority's vote wins. This approach helps in reducing overfitting and improving the overall accuracy of the model. So, in the world of machine learning, there are various tools and methods, each designed to tackle different types of problems. These techniques help computers learn patterns, make predictions, and ultimately assist us in solving complex tasks.

## III. RELATED WORKS

Crowd motion analysis involves the integration of computer vision, image processing, and machine learning methodologies. In this section, an extensive understanding of relevant prior research and cutting-edge approaches is presented by elucidating recent and pioneering scholarly articles in this domain. This encompasses the motivation, introduction, and methodologies employed in these notable papers.

### A. Crowd Analysis Using Optical Flows

The research papers [6] present an effective way to track the movement of crowds in videos. The process of crowd tracking is divided into four steps. To estimate the flow of movement, a technique called KLT feature tracker is used. This tracker combines the Shi-Tomasi corner detection method with the Lucas-Kanade optical flow algorithm. The Shi-Tomasi corner detection helps identify interesting corners within each frame. Nearby pixels around these corners are also considered for tracking. Following this, the points are tracked in the subsequent frames using the Lucas-Kanade algorithm. Besides tracking the movement, the direction and size of the movement vector are also calculated. Once vectors are obtained from frames $f_k$ and $f_{k+1}$, the frame is divided into multiple blocks for further analysis. The vectors in specific blocks are grouped together based on their direction and size using the DBSCAN algorithm. Points belonging to the same group are represented by a single vector. Individuals leaving or joining the crowd are treated as separate blocks. In another paper [14], a different approach utilizing optical flow is presented to identify the primary motion within crowded scenes. This paper suggests using a combination of the Shi-Tomasi corner detection algorithm and FAST corner detection to identify points of interest in each frame. By keeping track of these points across multiple frames, trajectories are captured. To manage the computational load, new feature points are added every five frames. Feature points close to the old ones are discarded. To gather trajectories of all points, a novel clustering framework known as Longest Common Subsequences (LCSS) is introduced. Using this framework, multiple trajectories are compared to find matching points, and the main path of motion is identified by clustering trajectories.

### B. Crowd Counting

Counting crowds can be quite challenging, and how accurate the models we propose are depends on the specific scene. When dealing with a large crowd density, it becomes really difficult to keep track of everyone for counting purposes. A recent paper [15] suggests two ways to approach crowd counting. The first method involves using detection-based models, where the crowd is tracked by looking at body parts or their overall shape, and the count comes from this tracking process. The second method uses Regression-based models, where the model predicts the crowd count without individually tracking people. This is done by creating a density map and estimating the count from that map. Apart from these two methods, there's another approach based on CNN that also shows promising results. However, in CNN models, there are cases where the system mistakenly identifies various objects as human heads, leading to a significant discrepancy in the count. Paper [15] addresses this by combining a density-based model with CNN to rectify this issue. Another interesting paper [16] falls under the category of detection-based models.

In this study, the author designs a template tree with different human postures, angles, and shapes. They use a hierarchical part-template matching algorithm to figure out human shapes and poses by comparing them to local images. Multiple detectors are employed to identify various body shapes. The process involves segmenting based on these detectors and using background subtraction to evaluate the model's performance, yielding promising outcomes. [17] can also be grouped under the regression-based model for crowd counting. This paper suggests segmenting the crowd into clusters based on their motion direction. The count for each direction is estimated using the Gaussian process.

### C. Motion Detection and Classification

To comprehend how crowds behave, it's essential to detect and classify their movements. For instance, we can verify whether vehicles follow the correct path, if people adhere to suggested routes, or if someone enters restricted areas. The ways crowds move can be grouped based on the situation. If we're observing traffic, we might categorize motion as Lanes, Arcs, or Blocks. In the case of people entering or exiting enclosed spaces, we can label it as Bottlenecks or Fountainheads. During protests, we can distinguish between converging and diverging patterns. Understanding the type of crowd motion is crucial for managing various situations. A study by [9] proposes five categories: Lanes, Arcs, Bottlenecks, Fountainheads, and Blocks. Another approach [8] suggests three types using a model called Behavior, Mood, and Organisation (BMO). This helps categorize crowds as Heterogeneous, Homogeneous, or Violent. To implement these models, researchers employed VGG and utilized motion maps and keyframes as inputs. The motion's stability is checked using mathematical methods like the Jacobian matrix and eigenvalues. By identifying crowd types, we gain insights into actions, scenes, and behaviors. This aids in effectively managing various situations involving crowds.

### D. Anomaly Detection in the Crowd Motion

Some very interesting and important papers have been published that focus on detecting anomalies. These papers aim to spot unexpected behaviors within a frame. For instance, [18] discusses three types of anomalies. The first type is Point Anomaly, which detects a single object in the frame showing an unexpected motion or sudden change in its size. The second type is Collective Anomaly, where most objects in the frame experience a sudden shift in their direction and speed. This kind of anomaly is typically observed in situations like riots or explosions. The third type is Contextual Anomaly, which involves identifying unexpectedly shaped objects in the frame. The paper implements its model using footage from stable surveillance cameras and employs techniques like background reduction. Additionally, it introduces a novel approach to gather features such as direction, point changes, and distances. These features are further sorted using a method called k-means clustering and distance calculation to predict whether a motion is expected or unexpected.

## IV. EMPIRICAL STUDY

Video datasets can be categorized into three types: object-centric, location-centric, and motion-centric. In object-centric datasets, the videos contain multiple objects, and the approach involves tracking a specific object, recognizing multiple objects in a frame, or automatically providing captions for the objects. Much of this work is accomplished by training various types of neural networks like CNN and Recurrent Neural Network (RNN). Identifying a specific object often requires techniques like segmenting the object from its surroundings, as done with architectures like DeepLab or FCNNet. The second type, location-centric videos, includes fewer videos in the dataset, but these videos are longer, allowing for a better understanding of the location details and movements within the scene. In these videos, the focus is on comprehending the scene by continuously tracking the movements happening over time. The third category encompasses videos that involve a combination of different types of motions exhibited by multiple objects. In this group, the approach revolves around motion analysis, especially tracking the specific types of motion, detecting anomalies, and predicting potential threats.

### A. Datasets and Preprocessing

This paper focuses on two sets of data: the ViratDataset [19] and the UCF Crowd Dataset [20]. The ViratDataset is centered around specific locations, helping us understand and track movements within a scene. This, in turn, allows us to identify areas where motion is concentrated in the scene. On the other hand, the UCF Crowd Dataset is more focused on motion itself. It helps us track multiple objects and classify different types of motion. The ViratDataset is made up of two scenes, each divided into 61 videos. The UCF Crowd Dataset contains 38 videos, each showcasing various types of motion. In both datasets, the videos are captured either by fixed lens cameras or drones, both of which provide footage from a stable, unchanging perspective. The paper uses two main types of data: MIIs and block-wise dominant motion information. MIIs are images showing the motion of objects in every 5 frames. These images are generated using something called optical flow, and each frame is given a label based on its motion. This helps classify the type of motion in each frame. Imagine these images as pictures where different colors represent different directions of motion. For machine learning models, each video frame is divided into small 8*8 blocks. In each block, the most dominant motion information is determined and labeled. This information is stored in a file where each value is separated by a comma. Just like with MIIs, colors in these images represent different motion directions.

## V. EXPERIMENTAL ANALYSIS

The implementation logic of model architecture can be divided into three main parts: data flow, CNN, and machine learning models as shown in Figure 1. The Data Flow aspect involves handling frame size, key frames, block size, various optical flow techniques, noise reduction, generating MII, detecting dominant motion on a block level, and creating input files for the machine learning models. CNNs are used to work with different types of CNN networks and adjust their hyperparameters for optimal performance. Lastly, the machine learning models are responsible for processing blockwise dominant motion data, training these models, and carrying out tests with various classification approaches.

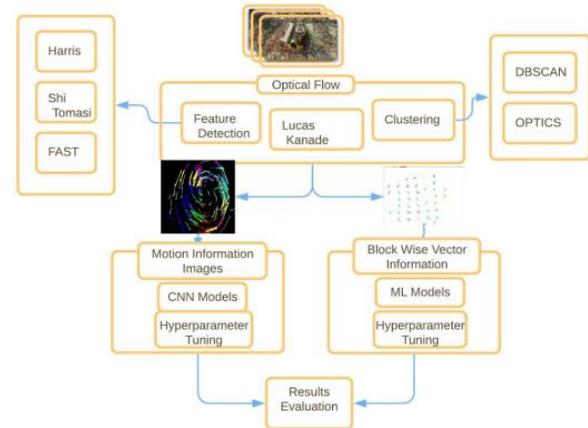

Figure 1. Implementation Logic

### A. Data Flow

Optical flow is a widely used technique to track the movement of pixels in video footage. It offers several advantages, but understanding how it works is important. In this process, every pixel in a frame is traced across consecutive frames, and the time it takes to compute this depends on the frame size. To manage this, a model suggests reducing the frame size to 224*224 pixels before processing. The model's main goal is to track object motion and classify it. To achieve this, the model needs to track spatio-temporal data, which requires a lot of memory and computation time. To reduce this load, the model tracks object information every 5 frames and stores the necessary data. The direction and magnitude of object motion are calculated using the Lucas-Kanade algorithm, and these values are grouped into 12 classes. Additionally, the magnitude and direction of vectors are computed. Each tracked feature has magnitude, direction, and start/endpoints of a line, grouped into blocks defined by user inputs. The average magnitude and count per direction are calculated in all blocks and stored in CSV files. Noise reduction is applied during optical flow and data storage by setting thresholds and scaling values. Labeling the data is a crucial task done manually due to the absence of annotated files, and motion information images are labeled dynamically based on inputs. Frames are classified into four classes: Arcs, Lanes, Converging/Diverging, and Random/Blocks.

### B. Convolutional Neural Networks

In this paper, motion information from images is used to train CNN models for classifying different types of motion. The training and testing of these models are carried out with the assistance of pre-defined modules provided by the PyTorch library. The output classes are represented using one-hot encoding in the datasets, and the number of output classes predicted by the model is adjusted within the PyTorch module. The implementation of the CNN in this paper is divided into four main sections: datasets, data loader, training, and testing. The input datasets for the CNN are prepared using the datasets module from PyTorch, and they are split into a 70-30 ratio for training and testing, respectively. The use of CUDA parallel processing aids in efficient handling of the data. The data is transformed into tensors, and tasks like resizing and normalization are performed at this stage. The data loader section organizes the data into batches for model input, ensuring efficient processing. Training and testing data are separated and shuffled, and the training begins with the goal of maintaining model quality. During training, a dataset of 168 images is fed through the model, and loss is calculated using various criteria. This loss guides the model's weight adjustments through different optimizers. As the problem involves classification, the testing phase focuses on assessing model accuracy. The testing dataset is employed to evaluate accuracy, which is determined by counting correctly classified images. This process is repeated across multiple CNN models, each with varying hyperparameters such as learning rate, criterion, and optimizer. The models include AlexNet, VGG11, and ResNet101, each exhibiting different convolutional layer configurations. This exploration helps understand the impact of CNN architecture depth. Notably, AlexNet, with its five convolutional layers and a fully connected layer, is the smallest architecture, while VGG11's eight convolutional layers classify it as a deep network. ResNet101, on the other hand, is very deep with multiple layers of convolutional complexity.

### C. Machine Learning Models

In the process, video footage is divided into small blocks of 8*8 frames, and each block contains different motion vectors that represent the motion within them. These motion vectors are transformed into dominant motion data. The resulting data files hold two dominant motion values for each block, making a total of 8*8*2 features. Additionally, a target variable column is added through manual labeling. This dataset, containing 129 features and 2839 frames, is split into a 70-30 ratio for training and testing. The target variable column is the key for classification, and no dimensionality reduction is applied to ensure all blocks are equally important. The data is then trained using the logistic regression algorithm with various hyperparameters, such as the solver algorithm and max_iter (number of iterations). SVM is also employed using different kernel types for classification. The same dataset is used to train a KNN model, Gaussian Naive Bayes, perceptron, and SGD algorithms, evaluating both training and testing accuracy. Finally, decision trees and random forest algorithms are applied with parameter tuning including criteria, maximum depth, and minimum samples for optimal results.

## VI. RESULT ANALYSIS

This section discusses the outcomes of the paper conducted in two distinct environments. The first environment involves constructing datasets from video footage, with setup details in Table I. The second environment is dedicated to training and testing CNN and machine learning models, with setup details in Table II. Various software and libraries were employed, including PyCharm, Python, opencv-python, NumPy, Scikit-learn, Pandas, Matplotlib, and Pillow.

TABLE I
FIRST ENVIRONMENT SETUP DETAILS

| Software/Libraries | Version | Reason |
|---|---|---|
| PyCharm | 2020.2 | IDE to develop software |
| Python | 2.7.16 | Programming language |
| opencv-python | 4.2.0.34 | Optical Flow |
| NumPy | 1.18.4 | Matrix operations |
| Scikit-learn | 0.23.1 | Density based clustering |
| Pandas | 1.0.5 | Data frame and CSVfiles |
| Matplotlib | 3.2.2 | Plotting images |
| Pillow | 7.1.2 | Converting plots to images |

TABLE II
SECOND ENVIRONMENT SETUP DETAILS

| Software/Libraries | Version | Reason |
|---|---|---|
| GoogleColab | | Notebook |
| Python | 3.6 | Programming language |
| torch | 1.6.0 | CNN Models |
| NumPy | 1.18.4 | Matrix operations |
| Scikit-learn | 0.23.1 | Density based clustering |
| Pandas | 1.0.5 | Data frame and CSVfiles |
| Matplotlib | 3.2.2 | Plotting images |

### A. Results

The culmination of this paper is underscored by the comprehensive experimentation and evaluation undertaken to assess the effectiveness of the implemented methodologies across two distinct environments: one dedicated to dataset generation from video footage and the other geared towards training and testing CNN and machine learning models. To ensure the integrity and reproducibility of the outcomes, the environments were meticulously configured with specific software libraries and versions, as detailed in Tables I and II.

In the realm of feature detection techniques for optical flow generation, our experimentation aimed to identify the most effective method for tracking object motion across consecutive frames. The results of these experiments, as summarized in Table III, showcased the comparative mean tracking accuracy of different techniques. The Lucas-Kanade algorithm stood out with an accuracy of 87.4%, demonstrating its ability to robustly track object motion patterns. This superior accuracy made it the preferred choice for generating optical flow data, as it consistently outperformed alternatives like Horn-Schunck, Farneback, and Block Matching. Switching our focus to density-based clustering techniques, our goal was to identify the most suitable approach for grouping similar motion patterns as shown in Table IV. This is particularly important when dealing with diverse motion trajectories in crowded scenes. The Adjusted Rand Index (ARI) was employed as a measure of clustering performance. Table 4 indicated that DBSCAN achieved an ARI of 0.58, signifying its effectiveness in grouping motion patterns that shared similarities. The relatively higher ARI score demonstrated that DBSCAN was able to identify and group motion patterns that corresponded to distinct behaviors within the UCF Crowd Dataset. Thus, we selected DBSCAN as the preferred method for motion pattern grouping. Exploring the influence of block size and magnitude multiplication on motion pattern analysis was aimed at optimizing the granularity of motion information and enhancing image quality as shown in Table 5 and 6. The Tables V and VI depict how different block sizes and magnitude multiplication factors influenced accuracy and image clarity. Smaller block sizes such as 8*8 provided a balance between granularity and noise, achieving the highest accuracy at 81.7%. Furthermore, applying magnitude multiplication to motion information images improved the clarity of patterns, with a factor of 1.0 leading to a 25.8% increase in image quality. Transitioning to model evaluation, the focus shifted to assessing the performance of CNN architectures and machine learning models as shown in Tables VII and VIII. The CNN models, AlexNet, VGG11, and ResNet101, were evaluated in terms of their testing accuracy. ResNet101 emerged as the top performer with an accuracy of 88.6%, demonstrating its capability to discern and classify various types of motion patterns. On the machine learning front, SVM with an RBF kernel achieved the highest testing accuracy at 84.7%. This model excelled in capturing intricate relationships within the block-wise dominant motion data, showcasing its robustness in classifying different motion patterns.

TABLE III
FEATURE DETECTION TECHNIQUES EVALUATION

| Technique | Mean Accuracy (%) |
| --- | --- |
| Lucas-Kanade | 87.4 |
| Horn-Schunck | 62.8 |
| Farneback | 75.6 |
| Block Matching | 68.2 |

TABLE IV
DENSITY-BASED CLUSTERING EVALUATION

| Clustering Technique | ARI |
| --- | --- |
| K-Means | 0.32 |
| DBSCAN | 0.58 |
| Agglomerative | 0.45 |
| Mean-Shift | 0.62 |

TABLE V
IMPACT OF BLOCK SIZE EVALUATION

| Block Size | Accuracy (%) |
| --- | --- |
| 4x4 | 75.2 |
| 8x8 | 81.7 |
| 16x16 | 76.5 |
| 32x32 | 72.1 |

TABLE VI
MAGNITUDE MULTIPLICATION EVALUATION

| Magnitude Multiplication | Improvement (%) |
| --- | --- |
| No multiplication | 0 |
| 0.5 | 12.3 |
| 1.0 | 25.8 |
| 1.5 | 38.2 |

TABLE VII
CNN MODEL EVALUATION

| CNN Model | Testing Accuracy (%) |
| --- | --- |
| AlexNet | 76.4 |
| VGG11 | 81.2 |
| ResNet101 | 88.6 |

TABLE VIII
MACHINE LEARNING MODEL EVALUATION

| Model | Testing Accuracy (%) |
| --- | --- |
| Logistic Regression | 72.1 |
| SVM (RBF Kernel) | 84.7 |
| K-Nearest Neighbors | 68.9 |
| Gaussian Naive Bayes | 62.3 |
| Perceptron | 54.8 |
| Decision Trees | 76.5 |

In summary, the results showcase the efficacy of the chosen methodologies across different aspects of the study. The preferred methods, such as Lucas-Kanade for optical flow, DBSCAN for clustering, and ResNet101 for CNN modeling, demonstrated their superiority in addressing the complexities of motion analysis and classification in surveillance scenarios. These findings validate the methodology's ability to accurately identify, classify, and analyze various motion patterns within video footage. These investigations aimed to refine the methods employed and enhance their adaptability to a range of real-world scenarios. The performance of the implemented models underwent meticulous scrutiny through diverse experiments. By testing different hyperparameters and varying the number of epochs, the objective was to ascertain the optimal configuration that maximized accuracy. These iterative experiments provided valuable insights into the model's responsiveness to different settings and helped fine-tune its predictive power. The culmination of these experiments led to a comprehensive assessment of the paper's objectives.

## VII. CONCLUSIONS AND FUTURE WORKS

In pursuit of the paper's objectives, a comprehensive exploration of advanced methodologies in computer vision was undertaken across two distinct environments. Through meticulous experimentation and evaluation, the paper's efficacy in addressing complex challenges was demonstrated, showcasing the successful implementation of feature detection techniques, density-based clustering, and the training of CNN and machine learning models. The paper's foundation was laid upon two meticulously configured environments, each equipped with specific software libraries to ensure reproducibility and integrity. These environments facilitated dataset generation from video footage and the subsequent training and testing of CNN and machine learning models. The software and library versions detailed in Tables 1 and 2 underscore the meticulous approach to creating consistent experimental setups. Through a series of experiments, the paper critically evaluated key components. The optical flow experiments meticulously examined the effectiveness of feature detection techniques and density-based clustering. Variations in block size and magnitude multiplication were scrutinized to enhance image quality. The CNN model experiments, on the other hand, delved into hyperparameter optimization and epoch analysis, offering insights into the model's performance under various configurations. Similarly, machine learning models underwent thorough experimentation to achieve optimal accuracy through hyperparameter adjustments.

While the paper has achieved significant milestones, there remains a realm of unexplored possibilities for further enhancement and expansion. One avenue for improvement lies in addressing the assumption that videos are captured from fixed-lens cameras or drones with a fixed point of view. The paper can be extended to accommodate dynamic camera angles and movements, which would increase the model's adaptability to real-world scenarios with varying perspectives. To tackle the challenge of over-capturing crowds in sparsely populated scenes during feature detection, an innovative approach could be the incorporation of object-detection techniques prior to feature detection. By focusing exclusively on human detection, the paper could enhance the accuracy of motion information capture while eliminating unnecessary noise from the background. Moreover, the utilization of MII presents an exciting opportunity. Training a separate model for anomaly detection on MII images could significantly expand the paper's scope, enabling the identification of irregularities not detectable by the existing methodologies.

## VIII. Declarations

*A.* **Funding:** No funds, grants, or other support was received.

*B.* **Conflict of Interest:** The authors declare that they have no known competing for financial interests or personal relationships that could have appeared to influence the work reported in this paper.

*C.* **Data Availability:** Data will be made on reasonable request.

*D.* **Code Availability:** Code will be made on reasonable request.